\documentclass[conference]{IEEEtran}
\IEEEoverridecommandlockouts
\usepackage{cite}
\usepackage{amsmath,amssymb,amsfonts}
\usepackage{algorithmic}
\usepackage{graphicx}
\usepackage{textcomp}
\usepackage{xcolor}
\usepackage{titlesec}
\titlespacing{\section}{0pt}{\parskip}{-\parskip}

\usepackage[strict]{changepage}

\usepackage{graphicx}
\usepackage{subcaption}
\usepackage{multirow}
\usepackage{booktabs}  
\usepackage{amsmath}
\usepackage{enumitem}
\usepackage{setspace}
\usepackage{array}
\usepackage{caption} 
\usepackage{amsmath}
\usepackage{amsfonts}
\usepackage{amssymb}

\title{TT-BLIP: Enhancing Fake News Detection Using BLIP and Tri-Transformer\\
{}
\thanks{}
}

\author{\IEEEauthorblockN{Eunjee Choi}
\IEEEauthorblockA{\textit{School of Electrical Engineering} \\
\textit{Korea University}\\
Seoul, South Korea\\
eun09ji@korea.ac.kr}
\and
\IEEEauthorblockN{Jong-Kook Kim}
\IEEEauthorblockA{\textit{School of Electrical Engineering} \\
\textit{Korea University}\\
Seoul, South Korea\\
Jongkook@korea.ac.kr}
}

\begin{document}
%
\maketitle

\begin{abstract}
 Detecting fake news has received a lot of attention. Many previous methods concatenate independently encoded unimodal data, ignoring the benefits of integrated multimodal information. Also, the absence of specialized feature extraction for text and images further limits these methods. This paper introduces an end-to-end model called TT-BLIP that applies the bootstrapping language-image pretraining for unified vision-language understanding and generation (BLIP) for three types of information: BERT and BLIP\textsubscript{Txt} for text, ResNet and BLIP\textsubscript{Img} for images, and bidirectional BLIP encoders for multimodal information. The Multimodal Tri-Transformer fuses tri-modal features using three types of multi-head attention mechanisms, ensuring integrated modalities for enhanced representations and improved multimodal data analysis. The experiments are performed using two fake news datasets, Weibo and Gossipcop. The results indicate TT-BLIP outperforms the state-of-the-art models. 
\end{abstract}

\section{Introduction}
\label{sec:intro}

The expansion of digital platforms has altered news consumption by broadening access and increasing exposure to misinformation, emphasizing the urgent need for enhanced fake news detection. Social media advancements have evolved news delivery from traditional text articles to multimedia narratives \cite{hangloo2022combating}, integrating images and videos. Images provide visual information \cite{jin2016novel} that enhances the textual content and can attract more attention.

Fig. 1 shows two examples from the Weibo dataset. The first image and its text are easily verified as true. However, the second image and text, if viewed separately, may provide insufficient information to assess the authenticity of the news content. By examining both the image and text together the fabrication becomes clear, illustrating how combining image and textual information improves fake news detection.

\begin{figure}
    \centering
    \begin{subfigure}[b]{0.45\linewidth}
        \centering
        \includegraphics[width=4cm,height=3cm]{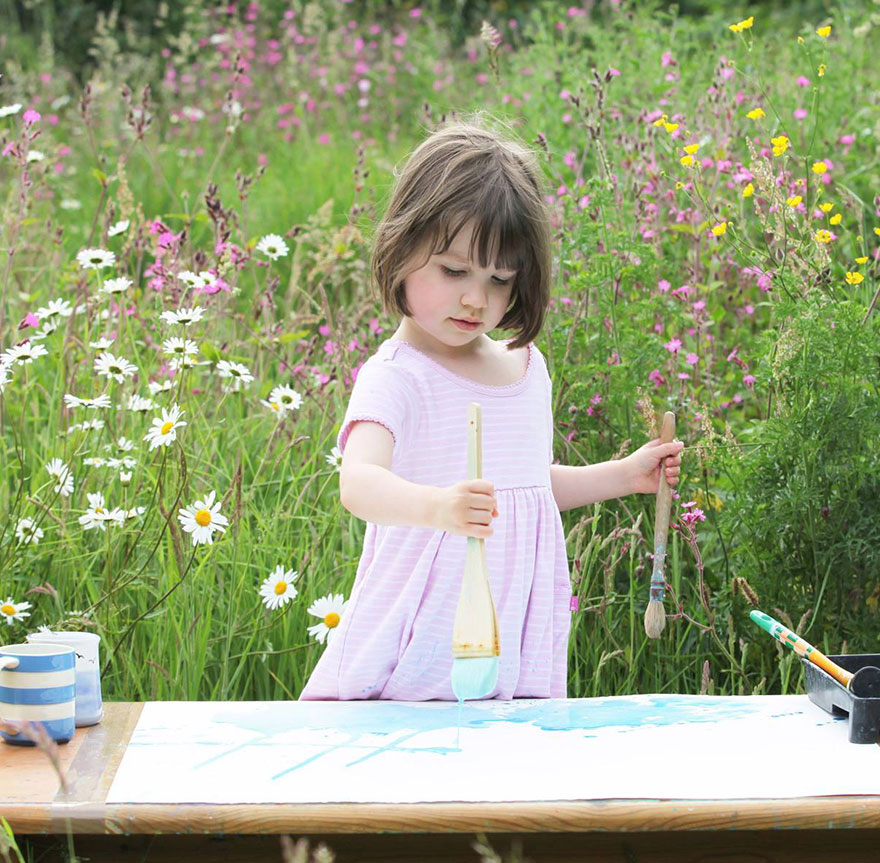} 
        \caption{Iris Grace, a British autistic girl with amazing painting talent.}
    \end{subfigure}
    ~
    \begin{subfigure}[b]{0.45\linewidth} 
        \centering
        \includegraphics[height=3cm]{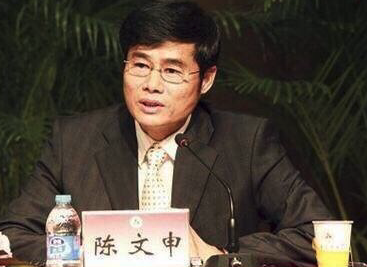}
        \caption{Eight party members and leading cadres were notified for disciplinary issues.}
    \end{subfigure}
    \caption{Real news (a) and fake news (b) examples from the Weibo dataset}
    \label{fig:example_images}
\end{figure}

The evolution of fake news detection techniques has transitioned from traditional approaches to more advanced deep learning methodologies. Previously, \cite{castillo2011information},\cite{conroy2015automatic} primarily utilized text content to discern between fake and real news. Building on this, Ma et al. \cite{ma2016detecting} delved into the potential of deep neural networks for representing tweets, emphasizing temporal-linguistic attributes. Chen et al. \cite{chen2018call} then augmented this approach by incorporating attention mechanisms into RNN structures. Deep learning-based fake news detection highlights notable enhancements in performance over their conventional counterparts, attributing this to the superior feature extraction capabilities of the newer methods. Jin et al. \cite{jin2016novel} adopted an end-to-end approach by merging image, textual, and social context features, using attention mechanisms to inform their predictions. Meanwhile, Wang et al. \cite{wang2018eann} introduced an event discriminator that aimed to identify applicable features across various events and thereby enhancing the model adaptability to novel events. Existing models have a clear deficiency in specialized feature extraction for both text and images and using cross modal attention has shown its own set of challenges. These two main challenges made the effective detection of fake news more difficult. 

This paper introduces Tri-Transformer BLIP (TT-BLIP) shown in Fig. 2, which is based on bootstrapping language-image pretraining for unified vision-language understanding and generation \cite{li2022blip} to address the challenges above. The proposed model uses BLIP \cite{li2022blip}, BERT \cite{devlin2018bert}, and ResNet \cite{he2016deep} for feature extraction. It is a three-pathway model employing BERT and BLIP\textsubscript{Txt} for text feature extraction, ResNet and BLIP\textsubscript{Img} for image feature extraction and a pair of bidirectional BLIP encoders for image-text correlation information. The Multimodal Tri-Transformer fuses features from text, image, and image-text modalities for fake news detection. The text modality employs self multi-head attention for internal analysis, while image and image-text modalities use cross-attention with text queries, aligning visual content with textual context. The process prioritizes text-driven analysis essential for evaluating multimodal fake news data. After the attention, each modality's output undergoes Multi-Layer Perceptron (MLP) transformation, ensuring data uniformity.  The processed outputs from different modalities are then concatenated, leading to a unified and comprehensive representation. Experiments are conducted using two multimodal fake news datasets to assess the performance of TT-BLIP: Weibo \cite{jin2017multimodal} and Gossipcop \cite{shu2020fakenewsnet}. The results demonstrate that TT-BLIP is the state-of-the-art model for fake news detection. \\

  In summary, the main contributions of this paper are:
  \begin{itemize}[topsep=0pt,itemsep=-1ex,partopsep=1ex,parsep=1ex]
   \item  TT-BLIP is proposed, which uses the pre-trained BLIP model for feature extraction to detect fake news.
   \item A novel fusion mechanism is introduced, which is a Multimodal Tri-Transformer. By fusing text, images, image-text features, the transformer captures the semantic information of all three modalities.
   \item The TT-BLIP is tested using two multimodal fake news, Weibo and Gossipcop, and results show that TT-BLIP peforms better than the state-of-the-art multimodal detection models.   
  \end{itemize}

\section{RELATED WORK}
In the domain of Multimodal fake news detection, various strategies have emerged that emphasize extracting features from both images and texts from news articles. EANN \cite{wang2018eann} is designed to detect fake news from social media, particularly those involving event-invariant features. MVAE \cite{khattar2019mvae} is an end-to-end network developed for fake news detection that incorporates a bimodal variational autoencoder and a binary classifier. Spotfake by Singhal et al. \cite{singhal2019spotfake} uses BERT \cite{devlin2018bert} for textual feature extraction and VGG19 \cite{simonyan2014very} for image features to efficiently detect fake news. They later refined their approach in Spotfake+ \cite{singhal2020spotfake+} to better identify fake news in full articles. SAFE \cite{zhou2020safe} explored the relationship between textual and image information in news articles for fake news detection. CAFE \cite{chen2022cross} detects fake news from social platforms by adapting to cross-modal ambiguities and analyzing uni-modal and cross-modal features. LIIMR \cite{singhal2022leveraging} selectively diminishes information from less significant modalities, while emphasizing and extracting related data from the dominant modality for each sample. MCAN \cite{wu2021multimodal} utilizes co-attention layers within its three sub-networks to merge textual and image features. This approach emphasizes capturing inter-dependencies across multimodal inputs for fake news detection. 

Some methods employ more information from datasets. DistilBert by Allein et al. \cite{allein2021like} utilize the latent representations of user-generated and shared content to detect disinformation in online news articles. By correlating user preferences and sharing behaviors, the model can efficiently differentiate between genuine and fake news without relying on user profiling during prediction. BDANN \cite{zhang2020bdann} tackles multimodal fake news detection on microblogging platforms by combining features from two modalities and eliminating event specific biases. FND-CLIP \cite{zhou2023multimodal} applied the Contrastive Language-Image Pre-Training (CLIP, \cite{radford2021learning}) vision-language pretraining model to measure the correlation between images and texts and utilize different modalities for decision-making. FND-CLIP used two unimodal encoders, two pair-wise CLIP encoders, guides the network learning for various modalities, and adeptly aggregates text, image, and fused features with a modality-wise attention module.  

Unlike previous models, TT-BLIP uses a three pathway model extracting features from image, text, and image-text multimodal information and the TT-BLIP model incorporates a Multimodal Tri-Transformer. This novel approach enables the fusion of text, images, and image-text features, offering a more comprehensive and effective method for utilizing multimodal information.

\section{METHOD}

\subsection{Overview}
In this paper, a new multimodal fake news detection model is proposed and Fig. 2 shows the overall architecture of TT-BLIP. TT-BLIP is composed of three modules: feature extraction module, feature fusion module, and fake news detector module. In the feature extraction module, image, text, and image-text features are extracted from text articles and associated images. In the fusion module, the MultiModal Tri-Transformer method is used to aggregate the output of the feature extraction module. Finally, in the fake news detector, the integrated features processed through the MultiModal Tri-Transformer are used to predict whether the content is real or fake.

\begin{figure*}[t]   
    \centering
    \setlength{\abovecaptionskip}{3pt}
    \setlength{\belowcaptionskip}{-3pt}
    \includegraphics[width=1.0\textwidth]{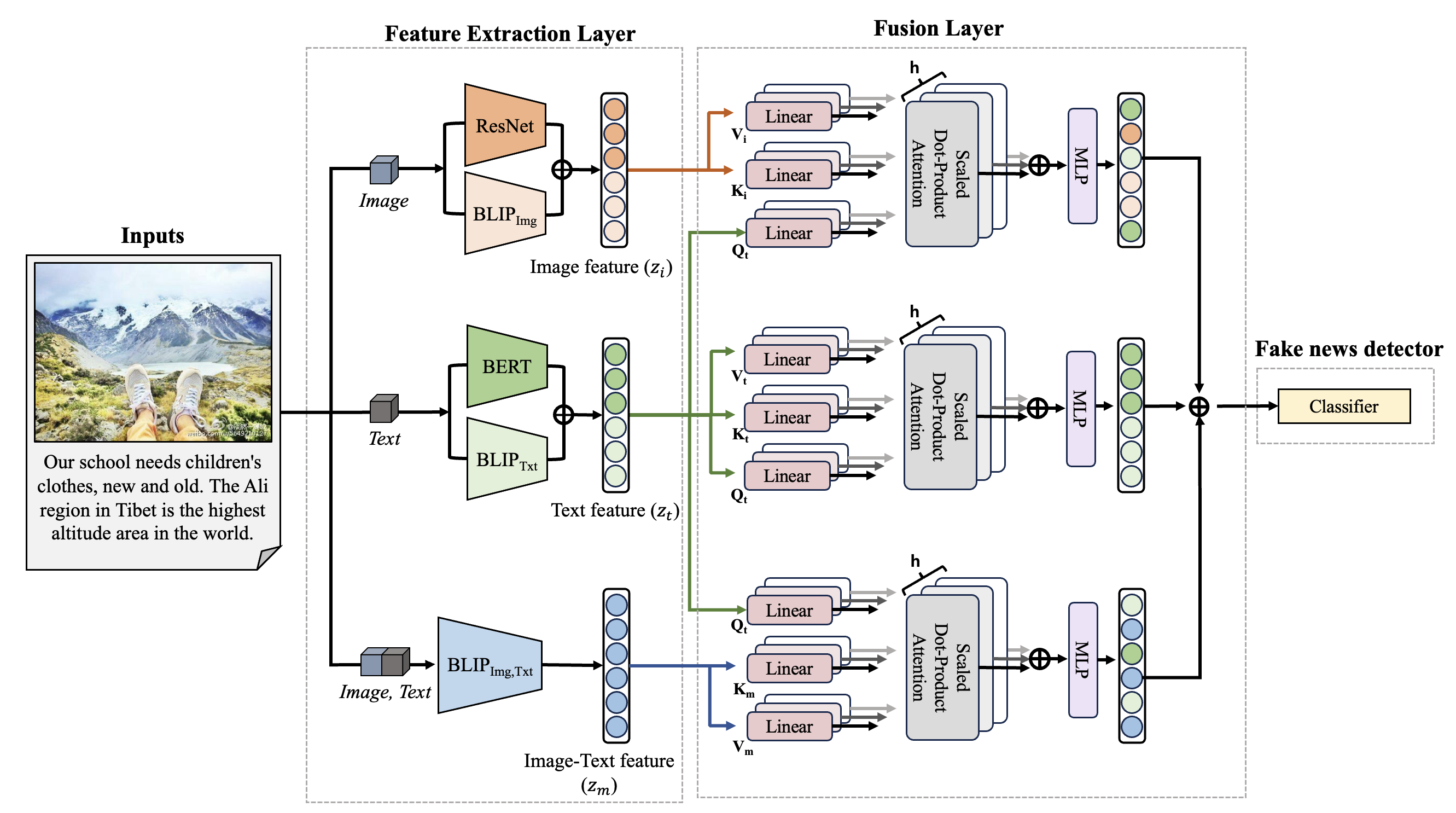}
    \caption{The architecture of the proposed TT-BLIP.}
    \label{fig:example_image}
\end{figure*}

\subsection{Feature Extraction Layer}
The feature extraction layer is comprised of text feature extraction layers, image feature extraction layers, and image-text feature extraction layers. The data used are denoted as \(x_{\text{Img}}\) for images and \(x_{\text{Txt}}\) for text.

\subsubsection{\textbf{Textual feature extractor}}: BERT \cite{devlin2018bert} and BLIP\textsubscript{Txt} \cite{li2022blip} are used in parallel to capture a comprehensive representation of the textual data.\\

\begin{itemize}[topsep=0pt,itemsep=-1ex,partopsep=1ex,parsep=1ex]
    
    \item \textbf{BERT-based Feature Extraction:} the pretrained BERT is used, bidirectional transformer model used for feature extraction. From this, the feature representation  \(z_{\text{t1}} = f_{\text{bert}}(x_{\text{Txt}})\) is obtained. The strength of BERT is its bidirectional training that enables it to understand words in their context and thus enhancing textual feature extraction. 

    \item \textbf{BLIP\textsubscript{Txt}-based Feature Extraction:} BLIP\textsubscript{Txt} is used to derive \(z_{\text{t2}} = f_{\text{blip\_T}}(x_{\text{Img}}, x_{\text{Txt}})\). BLIP is inherently designed to accept both text and images as inputs simultaneously. However, to ensure that only textual information is considered, an image tensor (dimension: $224 \times 224 \times 3$) filled with zeros is inserted for the image data. This ensures that BLIP focuses solely on the textual data and extracts relevant features accordingly.  
\end{itemize}

\subsubsection{\textbf{Image feature extractor}}

For the image data, ResNet \cite{he2016deep} and BLIP\textsubscript{Img} are used in parallel.\

    \begin{itemize}[topsep=0pt,itemsep=-1ex,partopsep=1ex,parsep=1ex]
    \item \textbf{ResNet-based Feature Extraction:} the pretrained ResNet model is employed for image feature extraction. The feature representation \(z_{\text{i1}} = f_{\text{resnet}}(x_{\text{Img}})\) is obtained. This choice ensures that our model can effectively capture intricate patterns and features from images.

    \item \textbf{BLIP\textsubscript{Img}-based Feature Extraction:} the BLIP\textsubscript{Img} is used to derive \(z_{\text{i2}} = f_{\text{blip\_i}}(x_{\text{Img}}, x_{\text{Txt}})\), which serves as an alternative approach for image feature extraction. BLIP is designed to handle both text and images as input simultaneously. However, to focus solely on extracting image features in this work a dummy text is input. This process ensures that the textual aspect does not influence the extraction of image features.      
    \end{itemize}

The two image and two textual feature representations are then integrated into single image and textual representations, respectively. For image features, this integrated representation is denoted as  \(z_{\text{i}}\), and for text features, it is denoted as \(z_{\text{t}}\). These integrated features are achieved by concatenating the two respective feature information shown as \(z_{\text{i}} = [z_{i1}; z_{i2}]\) for the image features and \(z_{\text{t}} = [z_{t1}; z_{t2}]\) for text features. These are then used in subsequent stages of the proposed model.

\subsubsection{\textbf{Image-text feature extractor}}: the pretrained BLIP model is used to extract image-text features, aiming to combine image and textual information. This process is essential to create a comprehensive representation that captures the interaction between image and text data. The resulting \text{Image-text feature} representation integrates information from both modalities and can be expressed as follows: \(z_{\text{m}} = f_{\text{blip}}(x_{\text{Img}}, x_{\text{Txt}} )\). Here, \(z_{\text{m}}\) represents the integrated feature representation.

\subsection{Feature Fusion Layer}

In the feature fusion layer (Fig 3), the MultiModal Tri-Transformer is introduced and this idea comes from MultiModal BiTransformer \cite{kiela2019supervised} that augments BERT embeddings with image-centric embeddings, facilitating enhanced representation learning. The Tri-Transformer uses three types of multi-attention\cite{vaswani2017attention} mechanisms. It applies cross-modal attention between text and both image-text and image modalities. This ensures the text modality, which is crucial for the task, is emphasized more than the others while keeping the image and image-text channels independent. For text alone, it employs self multi-head attention to enhance textual analysis.

\begin{equation}
\text{C\_Attention}_{i-t}(Q, K, V) = \text{softmax}\left(\frac{Q_t K_i^T}{\sqrt{d_h}}\right) V_i
\end{equation}

\begin{equation}
\text{C\_Attention}_{m-t}(Q, K, V) = \text{softmax}\left(\frac{Q_t K_m^T}{\sqrt{d_h}}\right) V_m
\end{equation}

\begin{equation}
\text{S\_Attention}_{t}(Q, K, V) = \text{softmax}\left(\frac{Q_t K_t^T}{\sqrt{d_h}}\right) V_t
\end{equation}

Let $Q_t$ represent the query vector for text modality, used to extract relevant information. The key vectors for the text, image, and image-text modalities are denoted by $K_t$, $K_i$, and $K_m$, determining data relevance. The value vectors for these modalities are represented by $V_t$, $V_i$, and $V_m$, holding the actual data to be retrieved. The $\text{softmax}(\cdot)$ function refers to the softmax function. The term $d_h$ indicates the dimensionality of each modality, and $T$ stands for transpose.

In the Transformer architecture, several attentions are computed in parallel, with each attention's output known as a head.  The $num^{th}$ head is computed as:

\begin{equation}
\text{head}^{num}_{*} = \text{Attention}_*\left(Q_t W^{Q_t}_{num},K_* W^{K_*}_{num}, V_* W^{V_*}_{num}\right)
\end{equation}

where $\mathbf{W}^{Qt}_{num} \in \mathbb{R}^{d_t \times d_q}$ is the weight matrix of $Q_t$ when computing the head of the $num^{th}$ text modality; $\mathbf{W}^{K*}_{num} \in \mathbb{R}^{d_* \times d_k}$ be the weight matrix of $K_*$ when computing the head of the $num^{th}$ $*$ modality; and $\mathbf{W}^{V*}_{num} \in \mathbb{R}^{d_* \times d_v}$ be the weight matrix of $V_*$ when computing the head of the $num^{th}$ $*$ modality, where $* \in \{i, t, m\}$.

Following this, all heads of the $*$ modalities are connected, denoted as $Y_*$, as follows:

\begin{equation}
\begin{split}
Y_* &= \text{MultiHead}(Q_t, K_*, V_*) \\
&= \text{Concat}(\text{head}^1_*, \text{head}^2_*, \ldots, \text{head}^n_*) \mathbf{W}_{O_*},
\end{split}
\end{equation}

Where $\mathbf{W}{O_*}$ is the weight matrix multiplied after splicing the head of $*$ modalities and $n$ denotes the number of self-attention heads used. Thus, the text-based image representation $f_{it}$, the text-based image-text representation $f_{mt}$, and text representation $f_{t}$ can be obtained as follows:

\begin{align}
f_{at} &= \text{MultiHead}(Y_i; \theta_{c-\text{att}}^i), \\
f_{vt} &= \text{MultiHead}(Y_m, \theta_{c-\text{att}}^m), \\
f_{t} &= \text{MultiHead}(Y_t, \theta_{\text{att}}^t)
\end{align}

where $\theta_{\text{att}}^i = \{\mathbf{W}_Q^i, \mathbf{W}_K^i, \mathbf{W}_V^i, \mathbf{W}_O^i\}$ and $\theta_{\text{att}}^m = \{\mathbf{W}_Q^m, \mathbf{W}_K^m, \mathbf{W}_V^m, \mathbf{W}_O^m\}$ represent the main hyperparameters required for the cross-attention module. $\theta_{\text{att}}^t = \{\mathbf{W}_Q^t, \mathbf{W}_K^t, \mathbf{W}_V^t, \mathbf{W}_O^t\}$ is used for the self-attention module.

After the multi-head attention step, each modality's output is passed through its own MLP (Multi Layer Perceptron) layer. The MLP layer perform linear transformations on data, integrating features extracted from previous layers to ensure a uniform format and facilitate the generation of the final output. The processed outputs are finally concatenated into a unified tensor. 

\subsection{Fake News Detector}
A fake news detector is used to verify the authenticity of multimodal news articles. The combined data from the fusion layer is sent to a classifier, which comprises of three linear layers with ReLU activations, batch normalization, and outputs binary classification for fake news detection. The label set is denoted by $\mathcal{Y}$. Specifically, a news article that is classified as 'fake' is assigned a label of 1, and 0 otherwise. To assess the difference between the predicted and actual labels, the cross-entropy loss is used and is formulated as:

\begin{equation}
L_{\text{cls}} = -\sum_{(p, y_i) \in (\mathcal{P}, \mathcal{Y})} \left[ y \cdot \log(\hat{y}_i) + (1 - y) \cdot \log(1 - \hat{y}_i) \right] 
\end{equation}

\begin{table*}[ht]
    \captionsetup{justification=centering}
    \centering
    \caption{Experimental Results on Weibo and Gossipcop Datasets\\ 
    \small {’-’ symbol indicates that the results are not available from the original paper.}}
    \small
    \begin{tabular}{llccccccc}  
        \toprule
        Dataset & Method & Accuracy & \multicolumn{3}{c}{Fake News} & \multicolumn{3}{c}{Real News} \\
        \cmidrule(lr){4-6} \cmidrule(lr){7-9}
        & & & Precision & Recall & F1 Score & Precision & Recall & F1 Score \\
        \midrule
        \multirow{12}{*}{Weibo} 
        & EANN \cite{wang2018eann} & 0.827 & 0.847 & 0.812 & 0.829 & 0.807 & 0.843 & 0.825 \\
        & MVAE \cite{khattar2019mvae} & 0.824 & 0.854 & 0.769 & 0.809 & 0.802 & 0.875 & 0.837 \\
        & Spotfake \cite{singhal2019spotfake} & 0.892 & 0.902 & \textbf{0.964} & 0.932 & 0.847 & 0.656 & 0.739 \\
        & SAFE \cite{zhou2020safe} & 0.762 & 0.831 & 0.724 & 0.774 & 0.695 & 0.811 & 0.748 \\
        & BDANN \cite{zhang2020bdann} & 0.821 & 0.790 & 0.610 & 0.690 & 0.830 & 0.920 & 0.870 \\
        & LIIMR \cite{singhal2022leveraging} & 0.900 & 0.882 & 0.823 & 0.847 & 0.908 & 0.941 & 0.925 \\
        & MCAN \cite{wu2021multimodal} & 0.899 & 0.913 & 0.889 & 0.901 & 0.884 & 0.909 & 0.897 \\
        & CAFE \cite{chen2022cross} & 0.840 & 0.855 & 0.830 & 0.842 & 0.825 & 0.851 & 0.837 \\
        & FND-CLIP \cite{zhou2023multimodal} & 0.907 & 0.914 & 0.901 & 0.908 & 0.914 & 0.901 & 0.907 \\
        & TT-BLIP(VGG) & 0.960 & 0.963 & 0.952 & 0.957 & \textbf{0.951} & 0.963 & 0.957 \\
        & TT-BLIP(XLNet) & 0.957 & 0.963 & 0.952 & 0.957 & \textbf{0.951} & 0.963 & 0.957 \\
        & \textbf{TT-BLIP(ours)} & \textbf{0.961} & \textbf{0.979} & 0.944 & \textbf{0.961} & 0.944 & \textbf{0.980} & \textbf{0.962} \\  
        \midrule
        \multirow{9}{*}{Gossipcop}
        & Spotfake+ \cite{singhal2020spotfake+} & 0.856 & - & - & - & - & - & - \\
        & LSTM-ATT \cite{lin2019detecting} & 0.842 & - & - & - & 0.839 & 0.842 & 0.821 \\
        & SAFE \cite{zhou2020safe} & 0.838 & 0.758 & 0.558 & 0.643 & 0.857 & 0.937 & 0.895 \\
        & DistilBert \cite{allein2021like} & 0.857 & 0.805 & 0.527 & 0.637 & 0.866 & \textbf{0.960} & 0.911 \\
        & CAFE \cite{chen2022cross} & 0.867 & 0.732 & 0.490 & 0.587 & 0.887 & 0.957 & 0.921 \\
        & FND-CLIP \cite{zhou2023multimodal} & 0.880 & 0.761 & 0.549 & 0.638 & 0.899 & 0.959 & 0.928 \\
        & TT-BLIP(VGG) & 0.846 & \textbf{0.842} & 0.552 & 0.667 & 0.847 & \textbf{0.960} & 0.900 \\
        & TT-BLIP(XLNet) & 0.865 & 0.655 & \textbf{0.792} & \textbf{0.717} & 0.875 & 0.933 & 0.903 \\
        & \textbf{TT-BLIP(ours)} & \textbf{0.885} & 0.737 & 0.596 & 0.659 & \textbf{0.910} & 0.950 & \textbf{0.930} \\
        \bottomrule
    \end{tabular}
    \label{tab:experimental_results}
\end{table*}

\section{EXPERIMENTS AND RESULTS}

\subsection{Dataset}
\subsubsection{Weibo}
The Weibo dataset was introduced by Jin et al. \cite{jin2017multimodal}. The dataset aggregates authentic news from esteemed Chinese news outlets, with the Xinhua News Agency being a prime example. The deceptive news articles were crawled and authenticated by Weibo’s official rumor verification system from May 2012 to January 2016. This system encourages general users to report potential misinformation, which is then assessed by a panel of reputable users. The experimental training set consists of 6,137 news articles, where there are 2,802 fake and 3,335 real news pieces, while the testing set comprises of 833 fake and 852 real news.

\subsubsection{Gossipcop}
Gossipcop dataset is used in this paper (\cite{shu2020fakenewsnet}). There are 10010 news articles in the training set, where 2036 are fake and 7974 are real news pieces. The testing set consists of 545 fake and 2285 real news.

\subsection{Experimental Settings} 
For text feature extraction, the  pretrained BERT model which is based on Chinese was used for the Weibo dataset, and the "bert-base-uncased" model for the Gossipcop dataset. In terms of image feature extraction, the input image size is set to 224 × 224 and ResNet \cite{he2016deep} was used. Lastly, image-text features are extracted by pairing images and text and processing them with the pretrained BLIP \cite{li2022blip} model. Because the BLIP is pretrained using English text, for the Weibo dataset the Google Translation API is utilized to translate Chinese texts into English. The model is trained using the Adam optimizer with default parameters, a learning rate of $1 \times 10^{-3}$ and a batch size of 64.    

\subsection{Results and Analysis}
TT-BLIP is compared to the state-of-the-art models which are presented in Table I to validate the performance. The experiments are performed using the Weibo and the Gossipcop datasets. For the evaluation, accuracy, precision, recall, and F1 scores are calculated for both real and fake news. Several strategies, such as EANN and Spotfake, use concatenation or attention mechanisms for feature fusion but often face correlation deficiencies because features remain in distinct semantic spaces. CAFE attempts to mitigate this through cross-modal alignment, aligning texts and images within a unified semantic space. Yet, its effectiveness is limited by small datasets and general labels, resulting in persistent semantic gaps between textual and image features. FND-CLIP uses two pre-trained CLIP encoders to extract the representations from the image and text. TT-BLIP achieved the highest accuracy of 96.1\% and 88.5\% in detecting fake news mainly due to the following reasons: 1) the TT-BLIP architecture combines ResNet and BLIP\textsubscript{Img} for image data and BERT and BLIP\textsubscript{Txt} for text processing, while using bidirectional BLIP encoders for correlation information. ResNet is used for extracting complex features from images, aiding in news image analysis. BERT is used for understanding language nuances, important for analyzing news text. Experiments with alternatives like TT-BLIP(VGG) (employing VGG  \cite{simonyan2014very} instead of ResNet) and TT-BLIP(XLNet) (using XLNet \cite{yang2019xlnet} instead of BERT) demonstrated the superiority of the original TT-BLIP setup. 2) The MultiModal Tri-Transformer uses attention mechanisms across multiple modalities, ensuring a comprehensive and integrated representation of the data. This approach allows for a more comprehensive understanding of context and improves feature extraction through multi-head attention analyzing data from various perspectives. By focusing on important text and merging features from all modalities, this architecture improves data representation, essential for better classification accuracy.

\begin{figure*}[ht]
  \centering
  \begin{subfigure}[b]{0.3\textwidth}
    \includegraphics[width=\textwidth]{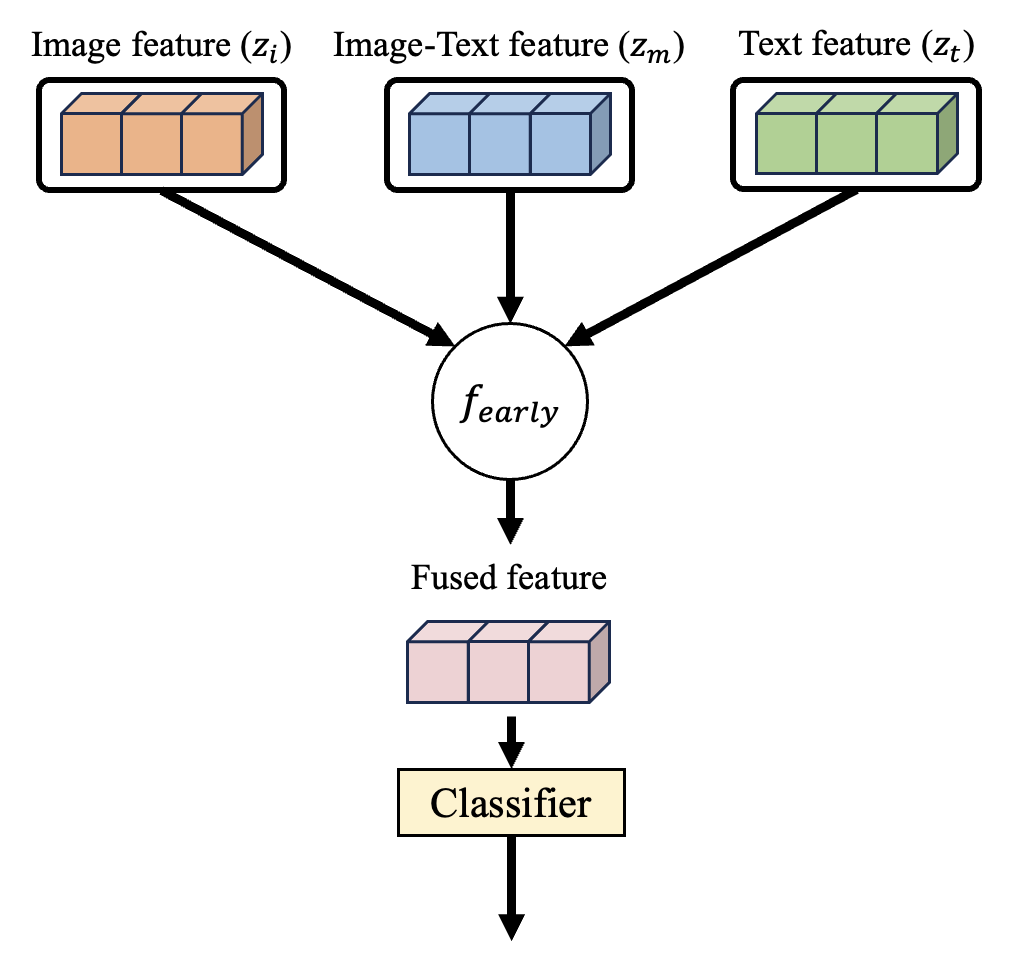}
    \caption{Early Fusion}
    \label{fig:image1}
  \end{subfigure}
  \hspace{2cm}
  \begin{subfigure}[b]{0.3\textwidth}
    \includegraphics[width=\textwidth]{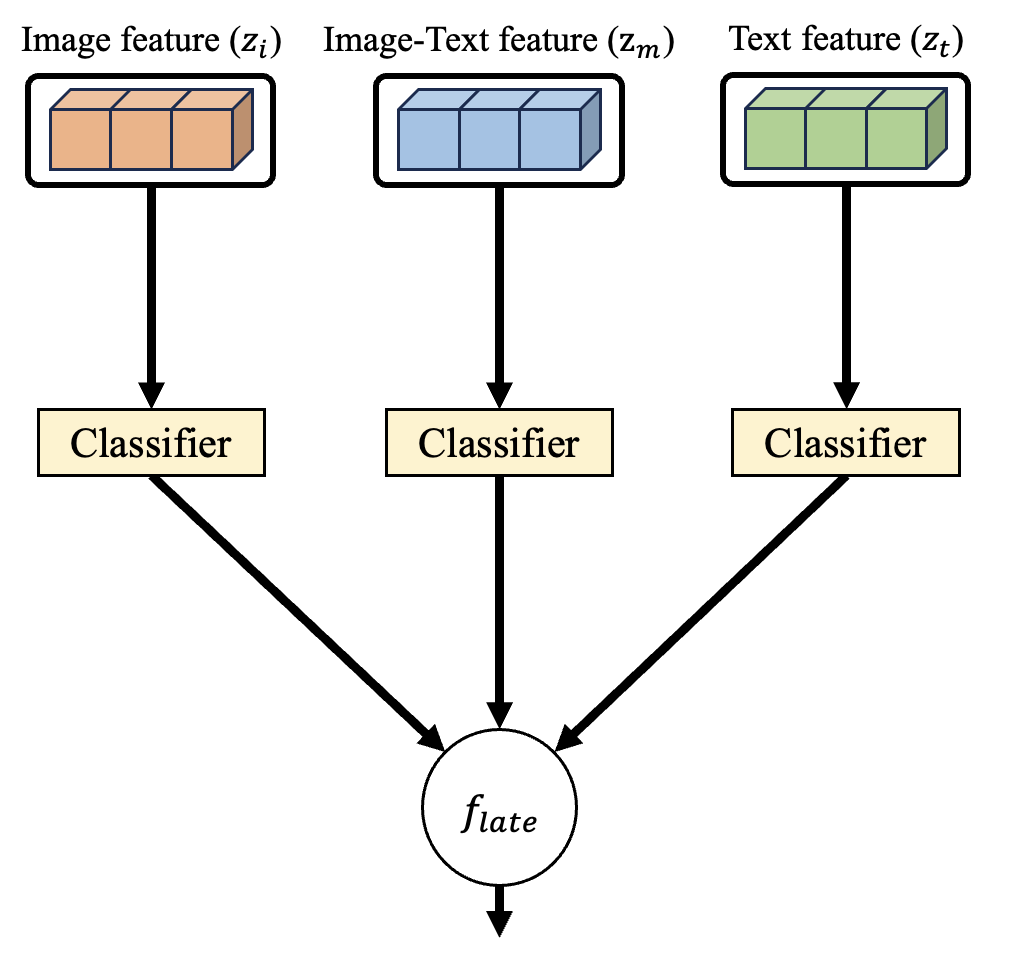}
    \caption{Late Fusion}
    \label{fig:image2}
  \end{subfigure}

  \vspace{2em} 

  \begin{subfigure}[b]{0.3\textwidth}
    \includegraphics[width=\textwidth]{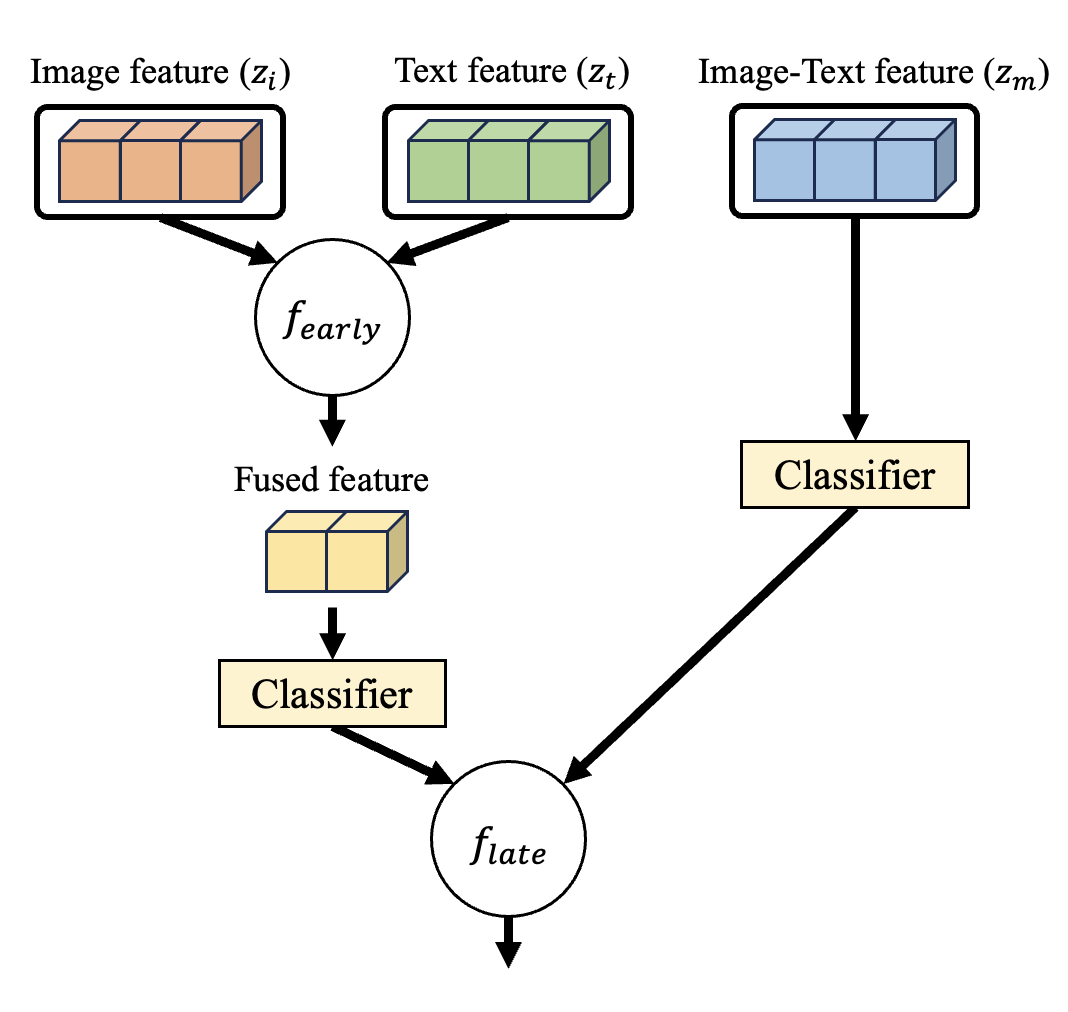}
    \caption{Hybrid Fusion}
    \label{fig:image3}
  \end{subfigure}
  \hspace{2cm}
  \begin{subfigure}[b]{0.333\textwidth}
    \includegraphics[width=\textwidth]{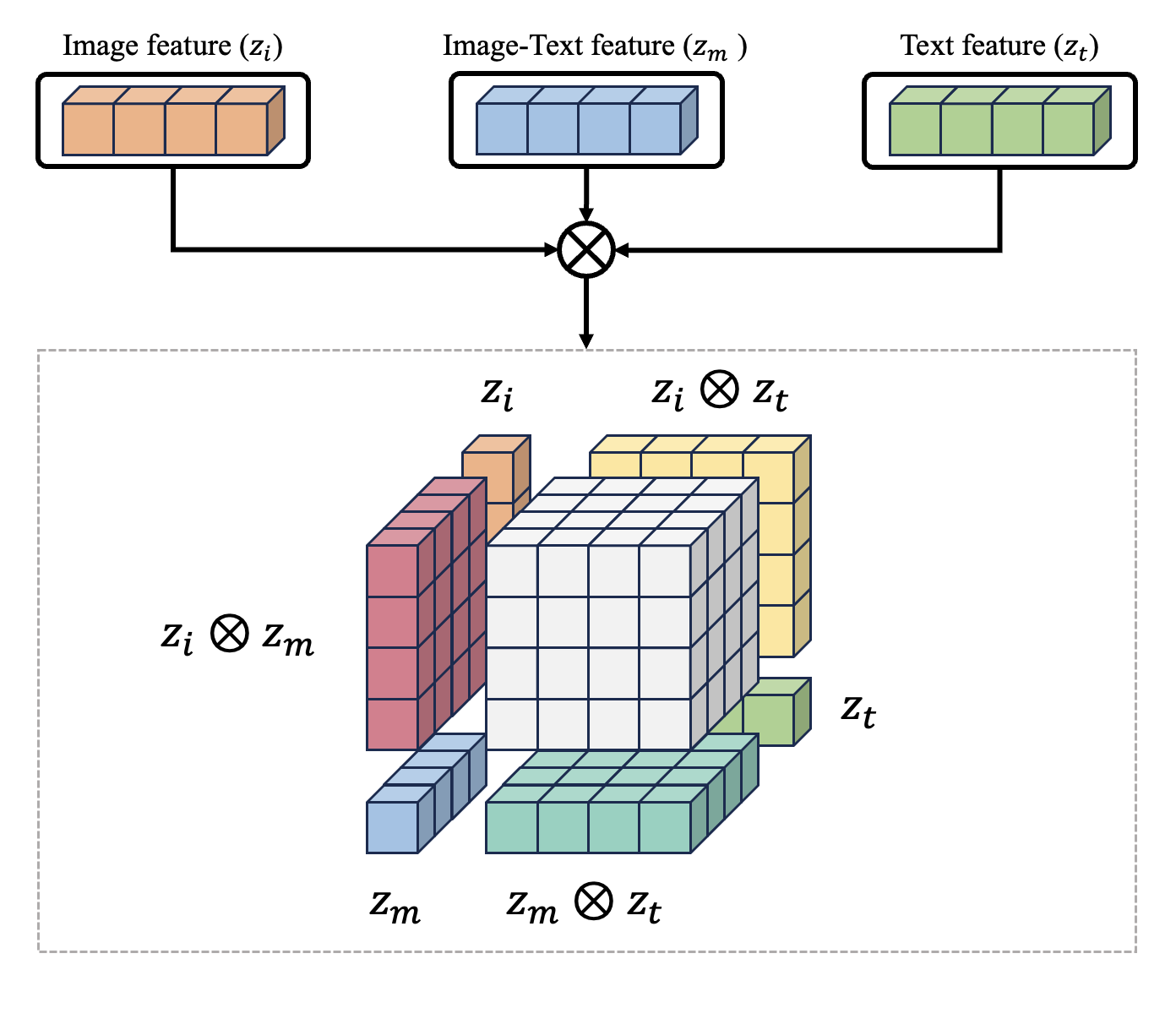}
    \caption{Tensor Fusion}
    \label{fig:image4}
  \end{subfigure}
  
  \caption{Architecture of different fusion strategies for Multimodal fake news detection}
  \label{fig:overall}
\end{figure*}

\subsection{Comparison of Fusion Methods}

\begin{table*}[ht]
    \captionsetup{justification=centering}
    \centering
    \caption{Comparative Analysis of Fusion Techniques for Fake News Detection on Weibo and Gossipcop Datasets}
    \small
    \begin{tabular}{llccccccc}  
        \toprule
        Dataset & Method & Accuracy & \multicolumn{3}{c}{Fake News} & \multicolumn{3}{c}{Real News} \\
        \cmidrule(lr){4-6} \cmidrule(lr){7-9}
        & & & Precision & Recall & F1 Score & Precision & Recall & F1 Score \\
        \midrule
        \multirow{5}{*}{Weibo} 
        & Early fusion \cite{snoek2005early} & 0.792 & 0.828 & 0.742 & 0.783 & 0.761 & 0.843 & 0.800 \\
        & Late fusion \cite{snoek2005early} & 0.806 & 0.840 & 0.761 & 0.799 & 0.777 & 0.852 & 0.813 \\
        & Hybrid fusion \cite{wu2005multi} & 0.815 & 0.850 & 0.770 & 0.808 & 0.785 & 0.861 & 0.821 \\
        & Tensor fusion \cite{zadeh2017tensor} & 0.821 & 0.876 & 0.786 & 0.829 & 0.766 & 0.864 & 0.812 \\       
        & \textbf{TT-BLIP} & \textbf{0.961} & \textbf{0.979} & \textbf{0.944} & \textbf{0.961} & \textbf{0.944} & \textbf{0.980} & \textbf{0.962} \\
        \midrule
        \multirow{5}{*}{Gossipcop}
        & Early fusion \cite{snoek2005early} & 0.634 & 0.277 & 0.560 & 0.370 & 0.861 & 0.651 & 0.742 \\
        & Late fusion \cite{snoek2005early} & 0.713 & 0.244 & 0.233 & 0.238 & 0.819 & 0.828 & 0.823 \\
        & Hybrid fusion \cite{wu2005multi} & 0.853 & 0.613 & \textbf{0.642} & 0.627 & \textbf{0.914} & 0.903 & 0.909 \\
        & Tensor fusion \cite{zadeh2017tensor} & 0.868 & \textbf{0.754} & 0.466 & 0.576 & 0.883 & \textbf{0.964} & 0.922 \\
        & \textbf{TT-BLIP} & \textbf{0.885} & 0.737 & 0.596 & \textbf{0.659} & 0.910 & 0.950 & \textbf{0.930} \\
        \bottomrule
    \end{tabular}
    \label{tab:experimental_results}
\end{table*}

\begin{figure}[htbp]
  \centering
  \begin{subfigure}{0.22\textwidth}
    \includegraphics[width=\textwidth]{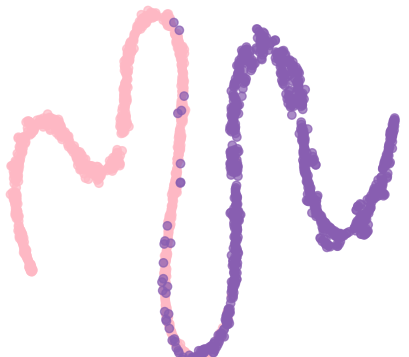}
    \caption{with text query}
    \label{fig:image1}
  \end{subfigure}
  \begin{subfigure}{0.22\textwidth}
    \includegraphics[width=\textwidth]{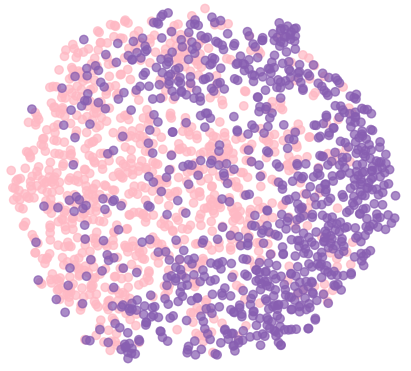}
    \caption{w/o fusion}
    \label{fig:image2}
  \end{subfigure}
  
  \hspace{1cm}
  
  \begin{subfigure}{0.22\textwidth}
    \includegraphics[width=\textwidth]{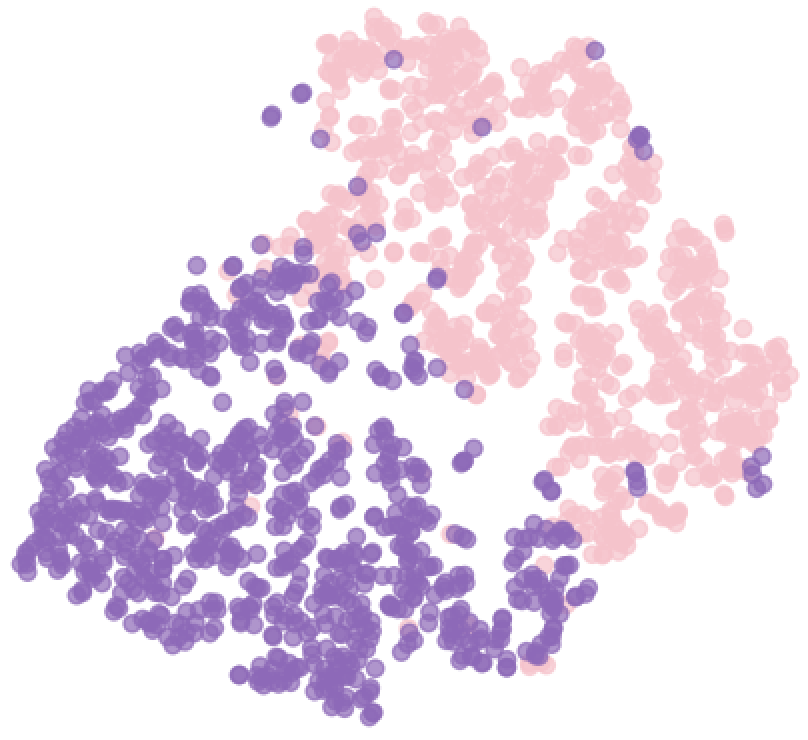}
    \caption{with image-text query}
    \label{fig:image3}
  \end{subfigure}
  \begin{subfigure}{0.22\textwidth}
    \includegraphics[width=\textwidth]{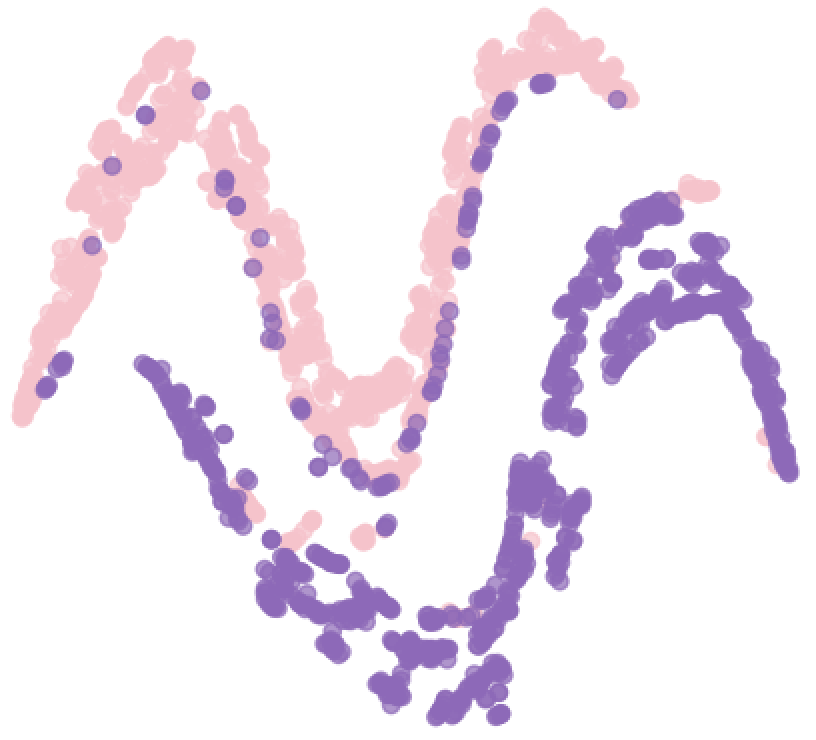}
    \caption{with image query}
    \label{fig:image4}
  \end{subfigure}
  \caption{t-SNE visualization of extracted features from the Weibo test set using TT-BLIP. Each color represents a distinct label grouping.}
  \label{fig:overall}
\end{figure}

In Table II, the TT-BLIP model, utilizing a multi-head attention mechanism, is compared against traditional fusion approaches such as early fusion \cite{snoek2005early}, late fusion \cite{snoek2005early}, hybrid fusion \cite{wu2005multi}, and tensor fusion \cite{zadeh2017tensor} on the Weibo and Gossipcop datasets. Early Fusion initially combines image, text, and image-text features, then applies classification to this unified dataset. Late Fusion aggregates results from different classifiers, each trained on separate modalities like image features, text features, and image-text features. Hybrid Fusion merges image, text, and image-text features by employing both early and late fusion methods. Tensor Fusion combines image, text, and image-text features using the outer product, enhancing interaction between modalities for improved classification. The architectures of traditional fusion methods are depicted in Fig. 3, demonstrating structural differences and highlighting the innovation of the TT-BLIP model.

In Table II, TT-BLIP outperforms traditional methods significantly. For the Weibo dataset, TT-BLIP significantly outperformed the traditional methods on Weibo with an accuracy of 96.1\% and achieved higher precision (97.9\%) and F1 scores (96.1\% and 96.2\%) for fake news detection. Other methods did not perform as well, with the highest accuracy among them being only 82.1\%. On the Gossipcop dataset, TT-BLIP was superior with 88.5\% accuracy, better than tensor fusion at 86.8\%. In terms of precision and recall for fake news detection, TT-BLIP performs competitively, though not leading, with figures at 73.7\% and 59.6\% respectively. This indicates that while TT-BLIP excels in overall accuracy and F1 score, it encounters competition from methods like Hybrid and Tensor Fusion in specific metrics.

In Fig. 4, t-SNE visualizations analyze features processed by TT-BLIP in different settings within the MultiModal Tri-Transformer's feature fusion mechanism, using the test set of Weibo. The visualizations compare before and after fusion, utilizing different types of queries: text, image-text, and image. The dots with the same color depicts data with same label. In Fig. 4, the visualizations show that before fusion (Fig. 4 (b)) data points are indicating a lack of clear separation between fake and real news instances. Applying TT-BLIP, there's a improvement in clustering, the real and fake news instances are now grouped more distinctly. This enhanced separation facilitates a more effective identification of fake news by allowing the model to better integrate the underlying patterns within the data. In Fig. 4, which displays various query configurations, a comparative analysis of (a), (c), and (d) shows differences in data clustering effectiveness. Specifically, configurations employing image-text and image queries, as shown in Fig. 4 (c) and (d), result in less defined separations between clusters. This outcome indicates that methods relying solely on image data are less effective for classification. The reduced clarity in clustering with image or image-text queries emphasizes the significant role of text in creating a clear and distinct feature space. Textual content offers crucial context, improving accuracy in differentiating categories, particularly in tasks like fake news detection. This comparison shows that incorporating textual information improves the model's classification efficacy.
 
The distinct clustering with text queries demonstrates the value of text in improving the model's capacity to distinguish between categories, essential for tasks such as fake news detection. Fig. 4 confirms the benefit of prioritizing text in multimodal feature fusion, yielding more distinct feature representations and supporting better classification outcomes.

\begin{table*}[ht]
    \captionsetup{justification=centering}
    \centering
    \caption{Ablation experimental results of TT-BLIP}
    \small
    \begin{tabular}{lccccccccc}  
        \toprule
        \multicolumn{1}{c}{\multirow{2}{*}{Dataset}} & \multicolumn{1}{c}{\multirow{2}{*}{ResNet}} & \multicolumn{1}{c}{\multirow{2}{*}{BLIP\textsubscript{Img}}} & \multicolumn{1}{c}{\multirow{2}{*}{BERT}} & \multicolumn{1}{c}{\multirow{2}{*}{BLIP\textsubscript{Txt}}} & \multicolumn{1}{c}{\multirow{2}{*}{BLIP\textsubscript{Img,Txt}}} & \multicolumn{1}{c}{\multirow{2}{*}{Fusion}} & \multicolumn{1}{c}{\multirow{2}{*}{Accuracy}} & \multicolumn{2}{c}{F1 Score} \\
        \cmidrule(lr){9-10} 
        & & & & & & & & Fake News & Real News \\
        \midrule
        \multirow{9}{*}{\centering Weibo} 
        & \checkmark & \checkmark & & & & & 0.659 & 0.684 & 0.629 \\
        & & & \checkmark & \checkmark & & & 0.837 & 0.836 & 0.839 \\
        & & \checkmark & & \checkmark & \checkmark & \checkmark & 0.858 & 0.846 & 0.869 \\
        & \checkmark & & \checkmark & & & \checkmark & 0.912 & 0.911 & 0.912 \\
        & & & & & \checkmark & & 0.747 & 0.747 & 0.747 \\
        & \checkmark & \checkmark & \checkmark & \checkmark & \checkmark & & 0.852 & 0.853 & 0.851 \\
        & & \checkmark & \checkmark & \checkmark & \checkmark & \checkmark & 0.959 & 0.959 & 0.959 \\
        & \checkmark & \checkmark & & \checkmark & \checkmark & \checkmark & 0.936 & 0.937 & 0.935 \\
        & \checkmark & \checkmark & \checkmark & \checkmark & \checkmark & \checkmark & \textbf{0.961} & \textbf{0.961} & \textbf{0.962} \\ 
        \midrule
        \multirow{9}{*}{\centering Gossipcop} 
        & \checkmark & \checkmark & & & & & 0.798 & 0.861 & 0.632 \\
        & & & \checkmark & \checkmark & & & 0.846 & 0.892 & 0.733 \\
        & & \checkmark & & \checkmark & \checkmark & \checkmark & 0.846 & \textbf{0.895} & 0.714 \\
        & \checkmark & & \checkmark & & & \checkmark & 0.848 & 0.715 & 0.902 \\
        & & & & & \checkmark & & 0.837 & 0.746 & 0.879 \\
        & \checkmark & \checkmark & \checkmark & \checkmark & \checkmark & & 0.837 & 0.886 & 0.712 \\
        & & \checkmark & \checkmark & \checkmark & \checkmark & \checkmark & 0.856 & 0.717 & 0.903 \\
        & \checkmark & \checkmark & & \checkmark & \checkmark & \checkmark & 0.846 & 0.692 & 0.897 \\
        & \checkmark & \checkmark & \checkmark & \checkmark & \checkmark & \checkmark & \textbf{0.885} & 0.659 & \textbf{0.930} \\
        \bottomrule
        \end{tabular}
    \label{tab:experimental_results}
\end{table*}

\subsection{Ablation study}

Some variants and components of the model were tested to identify the importance of the proposed model. The results are shown in Table III. If the fusion module was not used, the features were directly concatenated. The effectiveness of each components is measured through two metrics: Accuracy and F1 Score, with separate F1 Scores for Fake News and Real News.
The Text-only model, utilizing BERT and BLIP\textsubscript{Txt} for textual analysis, proved more effective than the Image-only model which employed ResNet and BLIP\textsubscript{Img} for image-based feature extraction. This indicates that textual features are more crucial in the classification of fake news. Integrating BLIP\textsubscript{Img}, BLIP\textsubscript{Txt}, and BLIP\textsubscript{Img,Txt} slightly outperformed the Text-only model, indicating the benefits of incorporating image features and textual data. The performance of the BLIP\textsubscript{Img,Txt} only model, which uses only BLIP\textsubscript{Img,Txt} combined feature without any fusion, did not reach the accuracy levels of the integrating BLIP\textsubscript{Img}, BLIP\textsubscript{Txt}, and BLIP\textsubscript{Img,Txt} method. This highlights the necessity for more integrated usage of features. The removal of the ResNet module resulted in a minor decline in performance, implying its lesser importance compared to other components. In contrast, the exclusion of the BERT component led to a significant reduction in effectiveness, emphasizing the critical role BERT plays in text processing. Models lacking fusion methods were less effective than TT-BLIP, underlining the importance of advanced fusion techniques. The complete TT-BLIP model, integrating ResNet, BLIP\textsubscript{Img}, BERT, BLIP\textsubscript{Txt}, BLIP\textsubscript{Img,Txt}, and fusion techniques, showed the highest accuracy in classifying fake news.

\section{Conclusion}

This study presented the Tri-Transformer BLIP (TT-BLIP), which uses the bootstrapping language-image pretraining (BLIP), that is used for enhanced vision-language understanding, and BERT for text feature extraction, BLIP and ResNet for image feature extraction, BLIP for image-text feature extraction and the Multimodal Tri-Transformer for the fusion of the feature from different modalities. 
A key contribution is the three-pathway structure of TT-BLIP, which processes text and images independently while also learning the correlation of the two different modals and the Multimodal Tri-Transformer that provides an efficient fusion mechanism to integrate information across these three modalities. The TT-BLIP outperformed the previous state of the art fake news detection models in accuracy by 5.4\% for the Weibo dataset (90.7\% versus 96.1\%) and 0.5\% for the Gossipcop (88\% versus 88.5\%).


\bibliographystyle{IEEEbib}
\bibliography{strings,refs}

\end{document}